\documentclass[runningheads]{llncs}

\usepackage[T1]{fontenc}
\usepackage{graphicx}
\usepackage{amsmath,amssymb}
\usepackage{booktabs}
\usepackage{multirow}
\usepackage{xcolor}
\usepackage{hyperref}
\hypersetup{hidelinks}

\makeatletter
\let\oldthebibliography\thebibliography
\let\endoldthebibliography\endthebibliography
\renewenvironment{thebibliography}[1]{%
  \oldthebibliography{#1}%
  \fontsize{8}{8}\selectfont%
  \setlength{\itemsep}{0pt}%
  \setlength{\parskip}{0pt}%
}{\endoldthebibliography}
\makeatother
\usepackage{algorithm}
\usepackage{algorithmic}
\usepackage{indentfirst}
\usepackage{orcidlink}

\graphicspath{{figures/main/}{figures/appendix/}{figures/}}

\begin{document}

\title{Unsupervised Multi-kernel Learning\\for Automated Algorithm Selection}
\titlerunning{Unsupervised Multi-kernel Learning for AAS}

\author{
Yihang Lu\inst{1,2}
\and Tome Eftimov\inst{3}\orcidlink{0000-0001-7330-1902}
\and Carola Doerr\inst{2}\orcidlink{0000-0002-4981-3227}
}
\authorrunning{Y. Lu et al.}

\institute{
Sorbonne Université, CNRS, INRAE, IRD, iEES, Paris, France\\
\and
LIP6, Sorbonne Université, CNRS, Paris, France\\
\and
Computer Systems Department, Jožef Stefan Institute, Ljubljana, Slovenia\\
}

\maketitle

\begin{abstract}
Automated algorithm selection in black-box optimization
typically relies on supervised models that map landscape features
to algorithm performance labels.
Such models are costly to train, benchmark-dependent,
and often fail to generalize to unseen problem classes.
We study an unsupervised alternative:
multi-kernel clustering over heterogeneous landscape representations,
in which problem instances are grouped without using performance labels in the clustering stage, and the resulting clusters are mapped post hoc to solver recommendations through a strictly separated three-stage evaluation protocol.
Drawing on two decades of advances in multiple kernel learning,
we adopt a multi-kernel $k$-means formulation
that jointly learns cluster assignments and kernel weights
over four heterogeneous landscape views:
ELA, DeepELA, DoE2Vec, and TransOptAS. On affine BBOB-derived selector tasks for Differential Evolution (DE) and
Particle Swarm Optimization (PSO) at a fixed evaluation budget, we report mean$\pm$standard-deviation selector profiles over 50 independent random seeds for stochastic configurations.
Multi-kernel clustering obtains the strongest mean profile on the DE portfolio and remains competitive with, and nominally ahead of, the leading baselines on the more compressed PSO portfolio, where differences among the best methods are small relative to stochastic variation.
In representative median-seed runs used for visualization, the learned kernel weights retain ELA and TransOptAS while assigning zero weight to DeepELA and DoE2Vec, providing a task-specific interpretation of which representations are retained by the multi-kernel model for selector-oriented grouping.

\end{abstract}

\section{Introduction}
\label{sec:intro}

Automated algorithm selection (AAS) is a central problem in black-box
optimization and a key component of modern AutoML systems
\cite{Kerschke2019Survey,HutterKV19}.
Given a problem instance, the goal is to identify, from a predefined
portfolio, the algorithm most likely to perform well.
The dominant approach casts AAS as a supervised learning problem:
landscape features extracted from the problem instance
are mapped to algorithm performance labels obtained from extensive
benchmarking \cite{Kostovska2023GECCO,KerschkeT19}.

Despite their empirical success on standard benchmarks,
supervised selectors face fundamental limitations.
Obtaining reliable performance labels requires exhaustive evaluation
across large algorithm--problem combinations,
incurring significant computational cost
\cite{Hansen2016COCO,TornedeWH20}.
More critically, the resulting models are inherently benchmark-dependent:
trained on specific suites such as BBOB \cite{bbobfunctions},
they tend to overfit to structural regularities of the training benchmark
and fail to generalize to unseen problem classes
\cite{Cenikj2025Wall,Cenikj2024CrossBenchmark}.
This lack of generalization has been documented
under leave-one-problem-out, problem-split,
and cross-benchmark evaluation protocols
\cite{Nikolikj2024CEC,Nikolikj2023Assessing},
where even state-of-the-art supervised selectors
often fail to outperform a simple single-best-solver baseline.

Recent studies further show that the difficulty is not confined
to the predictive model.
Different landscape feature representations
impose markedly different structural organizations
on the same problem instances,
leading to inconsistent similarity notions across views
\cite{Cenikj2025Wall}.
No single representation is known to robustly capture
algorithm suitability across problem classes,
which challenges the assumption underlying supervised AAS.

\medskip

Motivated by these observations, we propose an unsupervised alternative.
Rather than predicting performance labels,
we cluster optimization problem instances based solely on
their landscape characteristics,
and then map the resulting groups to solver recommendations post hoc.
This separates the label-free learning of the partition from the later use of performance data for cluster-to-solver routing and evaluation.
Unsupervised clustering has previously been applied
to analyze problem distributions and benchmark coverage
\cite{Skvorc2020Understanding,Lang2021ELABenchmark,Cenikj2022Selector}.
Cluster-based algorithm selection is also known from systems such as CSHC and SNNAP
\cite{Malitsky2013CSHC,Collautti2013SNNAP}; our focus differs by studying continuous black-box optimization with unsupervised multi-kernel integration of heterogeneous landscape views.

Modern AAS relies on multiple, heterogeneous landscape representations.
Classical Exploratory Landscape Analysis (ELA) features
\cite{Mersmann2011ELA}
capture low-level statistical and geometric properties,
while recent deep learning approaches---DoE2Vec \cite{VanStein2023Doe2Vec},
DeepELA \cite{Seiler2025DeepELA},
and TransOptAS \cite{Cenikj2024TransOptAS}---learn
transformation-aware embeddings from raw problem samples.
These representations are heterogeneous in both dimensionality and semantics,
making naive concatenation an unreliable integration strategy.

\medskip

Multiple kernel learning as a principled integration framework. Multiple kernel learning (MKL) provides a natural solution
to the challenge of integrating heterogeneous representations\cite{Yu2012MKKM,Lu2021Discrete}.
Over the past two decades, the MKL literature has produced
a rich family of methods that jointly learn cluster assignments
and kernel weights over a collection of base kernels,
without assuming commensurability across spaces
\cite{Lu2022Unified,Lu2022Scalable,Lu2022spl}.
State-of-the-art multi-kernel $k$-means (MKKM) methods
offer structured sparsity, hierarchical kernel selection,
and monotone descent properties that make them well-suited
for integration tasks with heterogeneous input spaces\cite{Lu2022Discrete,Lu2022Diverse,Jitao2022}.
We bring this body of work to the AAS setting,
treating each landscape view as a separate kernel source
and leveraging recent MKKM advances
to learn sparse, selector-oriented problem groupings.

\medskip

The central question we address is not merely whether problem instances
can be clustered, but whether the resulting structure
supports a useful post-hoc cluster-to-solver mapping
under EC-relevant evaluation criteria.
Our empirical study is therefore organized around selector quality:
clustering is learned in a strictly unsupervised manner,
and the clusters are converted to solver recommendations
only after training, using training-set performance statistics.
The resulting selectors are evaluated by gap closure and selector cost.

Our contributions are threefold:
(i) we introduce multi-kernel learning as a principled framework
for unsupervised algorithm selection from heterogeneous landscape views,
using a structured kernel integration strategy instead of naive concatenation;
(ii) we conduct an extensive selector-oriented evaluation
on DE and PSO portfolios over 8{,}280 affine BBOB-derived instances,
reporting mean$\pm$ standard-deviation profiles over 50 independent random seeds for stochastic configurations;
and (iii) we show that the learned kernel weights provide a task-specific and interpretable view of which representations are retained by the
multi-kernel model for selector-oriented grouping.

\section{Methodology}
\label{sec:method}

\subsection{Problem Formulation}
\label{sec:formulation}

Let $\mathcal{P} = \{p_1, \dots, p_n\}$ be a set of $n$ optimization
problem instances and $\mathcal{A} = \{a_1, \dots, a_r\}$ a portfolio
of $r$ algorithms.
We focus on a strictly unsupervised clustering setting:
no algorithm performance information is used to learn the clustering model.
Performance labels are used only after clustering, for post-hoc routing and held-out evaluation.

We formulate unsupervised AAS as a clustering problem:
partition $\mathcal{P}$ into $k$ groups,
each representing a \emph{behavioral regime}
of problems with similar landscape characteristics.
We treat $k$ as a clustering-granularity parameter rather than forcing it to equal the portfolio size. The value $k=8$ provides a moderate over-partition of the five-solver portfolios and is not interpreted as a one-to-one match between clusters and solvers. Multiple clusters are allowed to route to the same solver in the post-hoc mapping stage.
Formally, we seek a clustering function
$f\colon \mathcal{P} \to \{1, \dots, k\}$
based solely on problem landscape features.
The resulting cluster assignments are converted post hoc
to solver recommendations via a cluster-to-solver mapping
defined on the training partition.

Each problem instance $p_i$ is characterized through $m$ heterogeneous
feature views.
The $v$-th view is represented by
$\mathbf{X}^{(v)} \in \mathbb{R}^{n \times d_v}$, $v = 1,\dots,m$,
where $d_v$ is the view dimensionality.
We consider four heterogeneous landscape descriptions:
ELA, DeepELA, DoE2Vec, and TransOptAS.
All views are standardized independently to zero mean and unit variance.

Naive concatenation of all views into a single feature vector
implicitly assumes commensurability across heterogeneous spaces
and may distort the geometry of individual representations,
suppressing view-specific structure that is informative for selector-relevant
problem groupings.
A principled alternative is to represent each view as a separate kernel
and let the model learn their relative contributions. For each view $\mathbf{X}^{(v)}$, we construct a kernel matrix
$\mathbf{K}^{(v)} \in \mathbb{R}^{n \times n}$
encoding pairwise similarities in the $v$-th feature space
(linear kernels unless stated otherwise).
This yields a collection of $m$ base kernels
$\{\mathbf{K}^{(v)}\}_{v=1}^{m}$,
one per landscape representation.
The multi-kernel formulation preserves the geometric integrity
of each view while enabling the model to discover
which representations are most informative for solver-relevant grouping.

\subsection{Multi-Kernel \textit{k}-Means for Unsupervised AAS}
\label{sec:mkkm}

Multiple kernel $k$-means (MKKM) \cite{Yu2012MKKM,Lu2021Discrete,Lu2022Discrete}
is a mature and well-studied framework that jointly optimizes
cluster assignments and kernel weights over a collection of base kernels.
Over the past two decades, the MKL community has developed
a rich family of MKKM variants with structured sparsity,
hierarchical kernel selection,
and scalable optimization schemes
\cite{Lu2022Scalable}.

Let $\mathbf{H} \in \mathbb{R}^{n \times k}$ be the normalized cluster-indicator matrix, with $H_{ic}=|\mathcal{C}_c|^{-1/2}$ if instance $p_i$ belongs to cluster $c$ and $0$ otherwise, so that $\mathbf{H}^{\top}\mathbf{H}=\mathbf{I}_k$.
Let $\boldsymbol{\alpha} = (\alpha_1,\dots,\alpha_m)$ be a non-negative
view-weight vector, and for each view $v$,
let $\mathbf{w}_v = (w_{v1},\dots,w_{vq_v})$
be a non-negative within-view kernel weight vector. Each view may provide one or more normalized candidate kernels, with $q_v=1$ recovering the one-kernel-per-view case.
The combined kernel is
\begin{equation}
\label{eq:combined_kernel}
\mathbf{K}(\boldsymbol{\alpha}, \mathbf{W})
=
\sum_{v=1}^{m} \alpha_v
\sum_{\ell=1}^{q_v} w_{v\ell}\, \mathbf{K}^{(v,\ell)}.
\end{equation}

We minimize a kernel $k$-means objective augmented with
structured sparsity penalties:
\begin{equation}
\label{eq:objective}
\min_{\mathbf{H},\boldsymbol{\alpha},\mathbf{W}}
\;
\mathrm{Tr}\!\Bigl[
\mathbf{K}(\boldsymbol{\alpha},\mathbf{W})
\bigl(\mathbf{I} - \mathbf{H}\mathbf{H}^{\top}\bigr)
\Bigr]
+
\lambda\,\Omega_{\mathrm{view}}(\boldsymbol{\alpha})
+
\gamma \sum_{v=1}^{m} \alpha_v\,\Omega_{\mathrm{kernel}}(\mathbf{w}_v),
\end{equation}
subject to $\boldsymbol{\alpha}\geq 0$, $\mathbf{1}^{\top}\boldsymbol{\alpha}=1$ and $\mathbf{w}_v\geq 0$, $\mathbf{1}^{\top}\mathbf{w}_v=1$ for each $v$, together with cardinality constraints expressed through $\lVert\boldsymbol{\alpha}\rVert_0$ and $\lVert\mathbf{w}_v\rVert_0$ to enforce sparse but non-collapsed support. The first term is the standard within-cluster dispersion, while $\Omega_{\mathrm{view}}$ and $\Omega_{\mathrm{kernel}}$ are respectively view-level and within-view kernel-level redundancy-control terms. The support constraints prevent degenerate solutions in which the combined kernel collapses to a single view.

\subsection{Cluster-to-Solver Mapping and Evaluation}
\label{sec:evaluation}
 
The overall workflow proceeds in three strictly separated stages,
which we make explicit here to clarify the role of performance labels
and the train/test boundary.
 
\textbf{Stage~1: Unsupervised clustering.}
The clustering model is fitted on the training partition
using only landscape features.
No algorithm performance information is used at this stage.
The output is a clustering function
$f\colon \mathcal{P} \to \{1,\dots,k\}$
that assigns each problem instance to one of $k$ groups.
 
\textbf{Stage~2: Post-hoc cluster-to-solver mapping.}
Once the clustering is fixed, performance labels from the
training partition are used to assign a recommended solver to each cluster.
For each cluster $c$, let $\mathcal{I}_c^{\mathrm{tr}}$ denote
its training instances and $s_{ia}$ the normalized performance cost of solver $a$
on instance $i$; lower values are better.
We assign to cluster $c$ the solver with the best mean training cost:
\begin{equation}
a^{*}(c)
=
\arg\min_{a \in \mathcal{A}}
\frac{1}{|\mathcal{I}_c^{\mathrm{tr}}|}
\sum_{i \in \mathcal{I}_c^{\mathrm{tr}}} s_{ia}.
\label{eq:routing}
\end{equation}
Performance labels are used exclusively in this routing step, only after unsupervised training is complete.
 
\textbf{Stage~3: Selector evaluation.}
Each test instance is assigned to a cluster by the learned model
from Stage~1, then routed to the solver $a^{*}(c)$ from Stage~2.
All evaluation metrics are computed exclusively on the held-out
test partition $\mathcal{D}^{\mathrm{te}}$. Let $\hat{a}_i$ be the solver recommended for test instance $i$.
The \emph{selector cost} is
\begin{equation}
\mathrm{SC}=\frac{1}{|\mathcal{D}^{\mathrm{te}}|}\sum_{i\in \mathcal{D}^{\mathrm{te}}} s_{i\hat{a}_i}.
\label{eq:sc}
\end{equation}

\noindent The mean regret is defined as
\begin{equation}
\mathrm{MR}=\frac{1}{|\mathcal{D}^{\mathrm{te}}|}\sum_{i\in \mathcal{D}^{\mathrm{te}}}
\left(s_{i\hat{a}_i}-\min_{a\in \mathcal{A}} s_{ia}\right).
\label{eq:mr}
\end{equation}
which measures the average excess cost of the recommended solver
relative to the instance-wise best solver.
Lower values indicate better selector quality. 

The single best solver (SBS) and virtual best solver (VBS)
serve as lower and upper reference baselines,
both computed on the test partition:
\begin{equation}
\mathrm{SBS}=\min_{a\in \mathcal{A}}\frac{1}{|\mathcal{D}^{\mathrm{te}}|}\sum_{i\in \mathcal{D}^{\mathrm{te}}} s_{ia},
\qquad
\mathrm{VBS}=\frac{1}{|\mathcal{D}^{\mathrm{te}}|}\sum_{i\in \mathcal{D}^{\mathrm{te}}} \min_{a\in \mathcal{A}} s_{ia}.
\label{eq:sbs_vbs}
\end{equation}
\emph{Gap closure} is defined as
$\mathrm{GC} = (\mathrm{SBS} - \mathrm{SC}) / (\mathrm{SBS} - \mathrm{VBS})$,
with higher values indicating better selector quality.
SBS corresponds to $\mathrm{GC}=0$ (no improvement over the global
best fixed solver) and VBS corresponds to $\mathrm{GC}=1$
(perfect instance-level selection).
We report SC and MR alongside GC to keep both raw cost and absolute regret visible.

\section{Optimization}
\label{sec:optimization}
The objective in Eq.~\eqref{eq:objective} is non-convex jointly in
$(\mathbf{H}, \boldsymbol{\alpha}, \mathbf{W})$.
We adopt an alternating minimization strategy
that iterates between the following two steps.

\paragraph{Step 1: Update cluster assignments $\mathbf{H}$.}
With $(\boldsymbol{\alpha}, \mathbf{W})$ fixed,
the combined kernel $\mathbf{K}(\boldsymbol{\alpha},\mathbf{W})$
is fixed, and the objective reduces to a standard kernel $k$-means problem:
\begin{equation}
\min_{\mathbf{H}} \;
\mathrm{Tr}\!\bigl[\mathbf{K}(\mathbf{I} - \mathbf{H}\mathbf{H}^{\top})\bigr].
\label{eq:update_h}
\end{equation}
This is solved by applying kernel $k$-means on the combined kernel matrix,
which makes the objective monotonically non-increasing at each inner iteration
\cite{Yu2012MKKM}.

\paragraph{Step 2: Update kernel weights $(\boldsymbol{\alpha},\mathbf{W})$.}
With $\mathbf{H}$ fixed, define the per-kernel dispersion score
\begin{equation}
d_{v\ell}
= \mathrm{Tr}\!\bigl[\mathbf{K}^{(v,\ell)}(\mathbf{I} - \mathbf{H}\mathbf{H}^{\top})\bigr].
\label{eq:dispersion}
\end{equation}
The optimization over $(\boldsymbol{\alpha}, \mathbf{W})$ with fixed $\mathbf{H}$ is performed under the simplex and hard support constraints. Because the latter can make the feasible set non-convex, the update is treated as a support-constrained weight step \cite{Lu2021Discrete,Lu2022Discrete}. The accepted update leaves the objective unchanged or decreases it.

\paragraph{Convergence.}
The alternating updates therefore generate a monotonically non-increasing sequence of objective values, which is bounded below because the kernel matrices are positive semi-definite and $\mathrm{Tr}[\mathbf{K}(\mathbf{I}-\mathbf{H}\mathbf{H}^\top)]\geq 0$. With hard support constraints, the feasible set can be non-convex, so the convergence statement is limited to this monotonicity and termination, under the stated stopping criterion, at a fixed point of the alternating procedure.
The complete procedure is summarized in Algorithm~\ref{alg:mkkm}.

\vspace{-10pt}

\begin{algorithm}[htbp]
\caption{Multi-Kernel $k$-Means for Unsupervised AAS}
\label{alg:mkkm}
\begin{algorithmic}[1]
\REQUIRE Kernel matrices $\{\mathbf{K}^{(v,\ell)}\}$,
         number of clusters $k$,
         parameters $\lambda$, $\gamma$,
         maximum iterations $T$
\ENSURE  Cluster assignments $\mathbf{H}$,
         kernel weights $\boldsymbol{\alpha}$, $\mathbf{W}$
\STATE Initialize $\alpha_v = 1/m$ for all $v$;
       initialize $w_{v\ell}$ uniformly within each view;
       initialize $\mathbf{H}$ randomly
\FOR{$t = 1$ to $T$}
  \STATE Compute $\mathbf{K} \leftarrow \mathbf{K}(\boldsymbol{\alpha},\mathbf{W})$
  \STATE Update $\mathbf{H}$ by kernel $k$-means on $\mathbf{K}$
  \STATE Compute dispersion scores $d_{v\ell}$ for all $(v,\ell)$
  \STATE Update $(\boldsymbol{\alpha},\mathbf{W})$ by solving
         the support-constrained weight sub-problem in Step~2
  \IF{convergence criterion met}
    \STATE \textbf{break}
  \ENDIF
\ENDFOR
\RETURN $\mathbf{H}$, $\boldsymbol{\alpha}$, $\mathbf{W}$
\end{algorithmic}
\end{algorithm}

\vspace{-20pt}

\section{Experimental Study}
\label{sec:experiments}

\subsection{Experimental Protocol}
\label{sec:eval}

We study unsupervised algorithm selection on affine BBOB-derived problem
sets in dimension $d=10$
\cite{bbobfunctions,Dietrich2022Affine,Vermetten2023Affine},
following the recombination protocol of recent AAS generalization studies
\cite{Cenikj2025Wall,Cenikj2024CrossBenchmark}.
The benchmark comprises 552 problem pairs and 8{,}280 generated instances. Two algorithm portfolios are evaluated:
five DE variants \cite{Storn1997DE} and five PSO variants \cite{Kennedy1995PSO},
each treated as a separate selector task and denoted DE1--DE5 and PSO1--PSO5 in the benchmark performance matrix.
All Table~\ref{tab:main} comparisons are reported at budget $B=100$ with $k=8$. The value $k=8$ provides a moderate over-partition of the five-solver portfolios and is not interpreted as a one-to-one match between clusters and solvers.
We additionally include a $k$-sweep diagnostic over $k\in\{3,5,8,10,12\}$ in Section~\ref{sec:k_sweep} to assess how sensitive the selector utility is to this granularity choice.
Problem instances are represented through four landscape views:
ELA \cite{Mersmann2011ELA,Prager2024Pflacco},
DeepELA \cite{Seiler2025DeepELA},
DoE2Vec \cite{VanStein2023Doe2Vec},
and TransOptAS \cite{Cenikj2024TransOptAS}.
All views are standardized independently.
Stochastic configurations are evaluated with the same 50 independent random seeds across views and baselines, and Table~\ref{tab:main} reports mean$\pm$standard deviation.
Agglomerative clustering is deterministic under this protocol and is therefore reported with zero standard deviation.

We compare multi-kernel clustering against two families of baselines. \textbf{Single-view baselines} apply each of the four landscape views
in isolation with three clustering algorithms:
Gaussian Mixture Models~(GMM), $k$-Means, and Agglomerative Clustering,
yielding 12 configurations.
\textbf{Concatenation baseline} naively merges all four views
into a single feature vector and applies the same  algorithms.

Following the three-stage workflow defined in
Section~\ref{sec:evaluation},
clustering is trained on the training partition without labels,
the cluster-to-solver mapping is constructed from training-partition
performance statistics, and all selector metrics are evaluated
on the held-out test partition. The primary evaluation metrics are
Gap Closure~(GC,~$\uparrow$),
Mean Regret~(MR,~$\downarrow$),
and Selector Cost~(SC,~$\downarrow$).
SBS and VBS serve as fixed reference bounds:
for DE, $(\mathrm{SBS}, \mathrm{VBS}) = (0.1484,\,0.0664)$;
for PSO, $(\mathrm{SBS}, \mathrm{VBS}) = (0.1594,\,0.1034)$.

\subsection{Main Results}

Table~\ref{tab:main} reports the full selector profile
across all configurations.
Each row corresponds to a specific view--clustering combination.
The best result within each group is underlined;
the best mean per metric is in \textbf{bold}.

\begin{table*}[t]
\centering
\caption{%
  Selector profile on DE and PSO at the fixed evaluation budget and $k=8$.
  Stochastic configurations report mean$\pm$standard deviation over 50 independent random seeds; deterministic agglomerative rows have zero standard deviation.
  GC~$\uparrow$: Gap Closure. MR~$\downarrow$: Mean Regret. SC~$\downarrow$: Selector Cost.
  Best mean per metric is in \textbf{bold}; best mean within each group is underlined.
  Reference bounds---DE: SBS\,=\,0.1484, VBS\,=\,0.0664; PSO: SBS\,=\,0.1594, VBS\,=\,0.1034.
}
\label{tab:main}
\setlength{\tabcolsep}{3pt}
\resizebox{\textwidth}{!}{%
\begin{tabular}{llcccccc}
\toprule
& & \multicolumn{3}{c}{\textbf{DE}} & \multicolumn{3}{c}{\textbf{PSO}} \\
\cmidrule(lr){3-5}\cmidrule(lr){6-8}
\textbf{Group} & \textbf{Method}
  & GC~$\uparrow$ & MR~$\downarrow$ & SC~$\downarrow$
  & GC~$\uparrow$ & MR~$\downarrow$ & SC~$\downarrow$ \\
\midrule
\multirow{3}{*}{\shortstack{All Views\\(Concat)}}
  & GMM
    & $\underline{0.3562{\scriptstyle\pm 0.0075}}$ & $\underline{0.0528{\scriptstyle\pm 0.0006}}$ & $\underline{0.1192{\scriptstyle\pm 0.0006}}$
    & $0.0494{\scriptstyle\pm 0.0020}$ & $0.0533{\scriptstyle\pm 0.0001}$ & $0.1567{\scriptstyle\pm 0.0001}$ \\
  & $k$-Means
    & $0.3556{\scriptstyle\pm 0.0088}$ & $0.0528{\scriptstyle\pm 0.0007}$ & $0.1192{\scriptstyle\pm 0.0007}$
    & $0.0477{\scriptstyle\pm 0.0019}$ & $0.0534{\scriptstyle\pm 0.0001}$ & $0.1568{\scriptstyle\pm 0.0001}$ \\
  & Agg.
    & $0.3239{\scriptstyle\pm 0.0000}$ & $0.0554{\scriptstyle\pm 0.0000}$ & $0.1218{\scriptstyle\pm 0.0000}$
    & $\underline{0.0518{\scriptstyle\pm 0.0000}}$ & $\underline{0.0531{\scriptstyle\pm 0.0000}}$ & $\underline{0.1565{\scriptstyle\pm 0.0000}}$ \\
\midrule
\multirow{3}{*}{\shortstack{Single View\\(DeepELA)}}
  & GMM
    & $0.1879{\scriptstyle\pm 0.0400}$ & $0.0666{\scriptstyle\pm 0.0033}$ & $0.1330{\scriptstyle\pm 0.0033}$
    & $0.0481{\scriptstyle\pm 0.0111}$ & $0.0533{\scriptstyle\pm 0.0006}$ & $0.1567{\scriptstyle\pm 0.0006}$ \\
  & $k$-Means
    & $0.2719{\scriptstyle\pm 0.0131}$ & $0.0597{\scriptstyle\pm 0.0011}$ & $0.1261{\scriptstyle\pm 0.0011}$
    & $\underline{0.0486{\scriptstyle\pm 0.0061}}$ & $\underline{0.0533{\scriptstyle\pm 0.0003}}$ & $\underline{0.1567{\scriptstyle\pm 0.0003}}$ \\
  & Agg.
    & $\underline{0.2728{\scriptstyle\pm 0.0000}}$ & $\underline{0.0596{\scriptstyle\pm 0.0000}}$ & $\underline{0.1260{\scriptstyle\pm 0.0000}}$
    & $0.0409{\scriptstyle\pm 0.0000}$ & $0.0537{\scriptstyle\pm 0.0000}$ & $0.1571{\scriptstyle\pm 0.0000}$ \\
\midrule
\multirow{3}{*}{\shortstack{Single View\\(DoE2Vec)}}
  & GMM
    & $0.0002{\scriptstyle\pm 0.0009}$ & $0.0820{\scriptstyle\pm 0.0001}$ & $0.1484{\scriptstyle\pm 0.0001}$
    & $\underline{0.0225{\scriptstyle\pm 0.0062}}$ & $\underline{0.0548{\scriptstyle\pm 0.0003}}$ & $\underline{0.1582{\scriptstyle\pm 0.0003}}$ \\
  & $k$-Means
    & $\underline{0.0002{\scriptstyle\pm 0.0010}}$ & $\underline{0.0820{\scriptstyle\pm 0.0001}}$ & $\underline{0.1484{\scriptstyle\pm 0.0001}}$
    & $0.0201{\scriptstyle\pm 0.0052}$ & $0.0549{\scriptstyle\pm 0.0003}$ & $0.1583{\scriptstyle\pm 0.0003}$ \\
  & Agg.
    & $0.0000{\scriptstyle\pm 0.0000}$ & $0.0820{\scriptstyle\pm 0.0000}$ & $0.1484{\scriptstyle\pm 0.0000}$
    & $0.0077{\scriptstyle\pm 0.0000}$ & $0.0556{\scriptstyle\pm 0.0000}$ & $0.1590{\scriptstyle\pm 0.0000}$ \\
\midrule
\multirow{3}{*}{\shortstack{Single View\\(ELA)}}
  & GMM
    & $0.2960{\scriptstyle\pm 0.0320}$ & $0.0577{\scriptstyle\pm 0.0026}$ & $0.1241{\scriptstyle\pm 0.0026}$
    & $0.0436{\scriptstyle\pm 0.0028}$ & $0.0536{\scriptstyle\pm 0.0002}$ & $0.1570{\scriptstyle\pm 0.0002}$ \\
  & $k$-Means
    & $\underline{0.3253{\scriptstyle\pm 0.0100}}$ & $\underline{0.0553{\scriptstyle\pm 0.0008}}$ & $\underline{0.1217{\scriptstyle\pm 0.0008}}$
    & $0.0475{\scriptstyle\pm 0.0022}$ & $0.0534{\scriptstyle\pm 0.0001}$ & $0.1568{\scriptstyle\pm 0.0001}$ \\
  & Agg.
    & $0.3066{\scriptstyle\pm 0.0000}$ & $0.0569{\scriptstyle\pm 0.0000}$ & $0.1232{\scriptstyle\pm 0.0000}$
    & $\underline{0.0482{\scriptstyle\pm 0.0000}}$ & $\underline{0.0533{\scriptstyle\pm 0.0000}}$ & $\underline{0.1567{\scriptstyle\pm 0.0000}}$ \\
\midrule
\multirow{3}{*}{\shortstack{Single View\\(TransOptAS)}}
  & GMM
    & $\underline{0.0093{\scriptstyle\pm 0.0144}}$ & $\underline{0.0812{\scriptstyle\pm 0.0012}}$ & $\underline{0.1476{\scriptstyle\pm 0.0012}}$
    & $0.0316{\scriptstyle\pm 0.0037}$ & $0.0543{\scriptstyle\pm 0.0002}$ & $0.1577{\scriptstyle\pm 0.0002}$ \\
  & $k$-Means
    & $0.0009{\scriptstyle\pm 0.0026}$ & $0.0819{\scriptstyle\pm 0.0002}$ & $0.1483{\scriptstyle\pm 0.0002}$
    & $0.0315{\scriptstyle\pm 0.0037}$ & $0.0543{\scriptstyle\pm 0.0002}$ & $0.1577{\scriptstyle\pm 0.0002}$ \\
  & Agg.
    & $0.0000{\scriptstyle\pm 0.0000}$ & $0.0820{\scriptstyle\pm 0.0000}$ & $0.1484{\scriptstyle\pm 0.0000}$
    & $\underline{0.0322{\scriptstyle\pm 0.0000}}$ & $\underline{0.0542{\scriptstyle\pm 0.0000}}$ & $\underline{0.1576{\scriptstyle\pm 0.0000}}$ \\
\midrule
Multi-Kernel
  & \textbf{MKKM}
    & $\mathbf{0.3605{\scriptstyle\pm 0.0042}}$ & $\mathbf{0.0524{\scriptstyle\pm 0.0003}}$ & $\mathbf{0.1188{\scriptstyle\pm 0.0003}}$
    & $\mathbf{0.0521{\scriptstyle\pm 0.0032}}$ & $\mathbf{0.0531{\scriptstyle\pm 0.0002}}$ & $\mathbf{0.1565{\scriptstyle\pm 0.0002}}$ \\
\bottomrule
\end{tabular}%
}
\end{table*}

\paragraph{View heterogeneity has a material impact.}
The performance of single-view selectors varies substantially, confirming that the choice of landscape representation materially affects unsupervised selector utility.
On DE, the strongest single-view mean profile is obtained by ELA+$k$-Means (GC=$0.3253\pm0.0100$), whereas DoE2Vec and TransOptAS remain close to the SBS reference under most clustering choices.
On PSO, the single-view differences are much more compressed: the strongest single-view mean GC is DeepELA+$k$-Means ($0.0486\pm0.0061$), followed closely by ELA-based variants.
This spread supports the use of view-aware integration rather than assuming that one representation is uniformly reliable.

\paragraph{Naive concatenation is competitive but not uniformly decisive.}
At $k=8$, concatenation is a strong baseline on DE: Concat+GMM obtains GC=$0.3562\pm0.0075$ and Concat+$k$-Means obtains GC=$0.3556\pm0.0088$.
These values are higher than the strongest single-view DE baseline but remain slightly below MKKM.
On PSO, concatenation also lies in the leading group, with Concat+Agg. reaching GC=$0.0518$ and Concat+GMM reaching GC=$0.0494\pm0.0020$.
Thus, the results do not support treating concatenation as a weak baseline; rather, they show that structured multi-kernel integration provides a small but useful improvement over strong concatenation baselines in the $k=8$ setting.

\begin{figure}[htbp]
\centering
\includegraphics[width=0.5\linewidth]{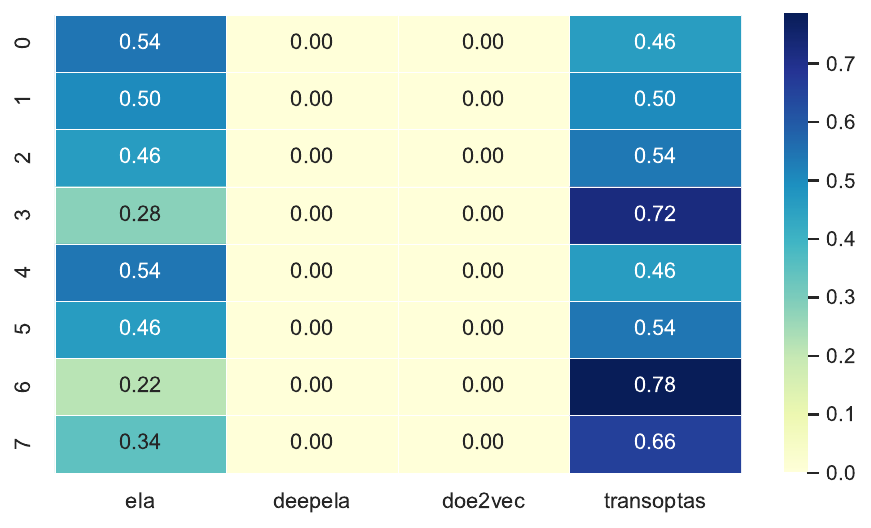}\hfill
\includegraphics[width=0.5\linewidth]{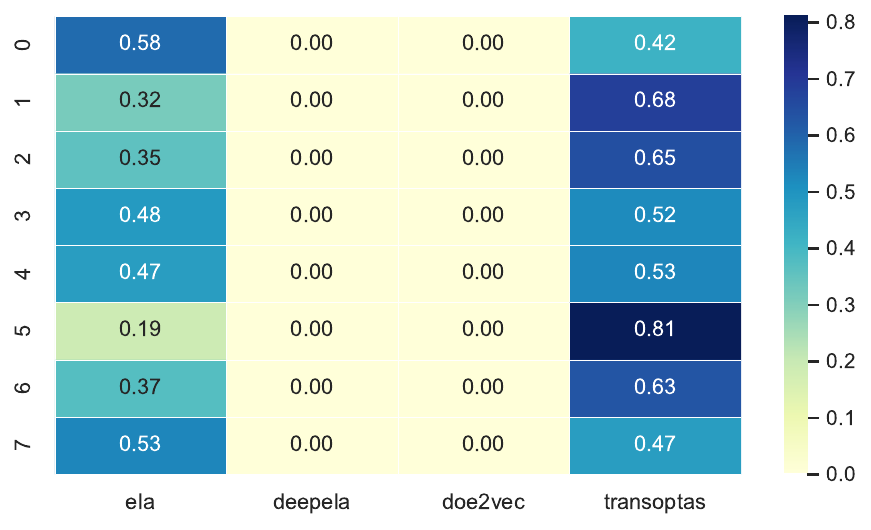}
\caption{%
Per-cluster normalized view contributions induced by multi-kernel clustering for DE (left) and PSO (right) in representative median-seed runs at $k=8$; the representative run is the seed whose test GC is closest to the median over the 50 seeds (DE seed 9, PSO seed 18).
DoE2Vec and DeepELA have zero normalized contribution in these runs.
ELA and TransOptAS carry the active normalized contribution, with cluster-specific and family-specific balance.%
}
\label{fig:sh_cluster_heatmaps}
\end{figure}

\paragraph{Multi-kernel clustering gives the strongest mean profile, with different strength on DE and PSO.}
On DE, MKKM obtains the strongest mean selector profile in Table~\ref{tab:main}, with GC=$0.3605\pm0.0042$, MR=$0.0524\pm0.0003$, and SC=$0.1188\pm0.0003$.
In relative GC terms, this corresponds to about $+10.8\%$ over the strongest single-view mean baseline (ELA+$k$-Means: $0.3253\pm0.0100$) and about $+1.2\%$ over the strongest concatenation mean baseline (Concat+GMM: $0.3562\pm0.0075$).
The latter gain is only about $0.0043$ GC, approximately one standard deviation of the MKKM GC values, so the percentage should be read as a marginal advantage whose interpretation is supported by the paired check below rather than as a large separation.
On PSO, MKKM is nominally ahead by mean GC, MR, and SC, reaching GC=$0.0521\pm0.0032$, MR=$0.0531\pm0.0002$, and SC=$0.1565\pm0.0002$.
However, the PSO margin over Concat+Agg. (GC=$0.0518$) is extremely small and lies well within the stochastic standard deviation of MKKM.
We therefore interpret the PSO result as competitive and mildly favorable, not as strong dominance.

\paragraph{Paired statistical check.}
Because the stochastic configurations use aligned seeds, we additionally compare GC values with two-sided Wilcoxon signed-rank tests.
The direction statements below refer to the observed mean GC differences, while the two-sided $p$-values assess whether the paired differences are nonzero.
On DE, MKKM has higher mean GC than the strongest stochastic concatenation baseline, Concat+GMM, and the paired difference is significant ($p=2.7\times10^{-4}$).
On PSO, MKKM has higher mean GC than the strongest stochastic concatenation baseline, Concat+GMM, and the paired difference is significant ($p=1.5\times10^{-5}$), but it is not significantly different from the deterministic Concat+Agg. reference when that singleton value is treated as a fixed comparator ($p=0.965$).
This supports the intended interpretation: a clear DE improvement, but only a competitive and non-dominant PSO result.

\paragraph{PSO is harder across the compared methods.}
All selectors achieve substantially lower GC on PSO than on DE.
This is consistent with prior observations that static landscape features have limited discriminative power for PSO-type portfolios \cite{Cenikj2025Wall}.
The reported mean$\pm$standard-deviation profiles reinforce this interpretation: PSO gains remain compressed even for the strongest methods, and small numerical differences should not be over-interpreted.

\subsection{Kernel Weight Analysis}
\label{sec:kernel_weights}

Figure~\ref{fig:sh_cluster_heatmaps} shows the per-cluster normalized view contributions induced by the multi-kernel model in representative median-seed runs at $k=8$. We define the representative run as the seed whose test GC is closest to the median over the 50 seeds, yielding seed 9 for DE and seed 18 for PSO.
In both algorithm families, DoE2Vec and DeepELA have zero contribution in these representative runs, while ELA and TransOptAS account for the active contribution.
This should not be read as evidence that DoE2Vec or DeepELA are generally uninformative for AAS; it only indicates that, under the present benchmark, fixed budget, and regularization regime, they do not add selector-relevant variation beyond the retained views.

\begin{figure}[p]
  \centering
  \vspace*{0.04\textheight}
  \includegraphics[height=0.80\textheight]{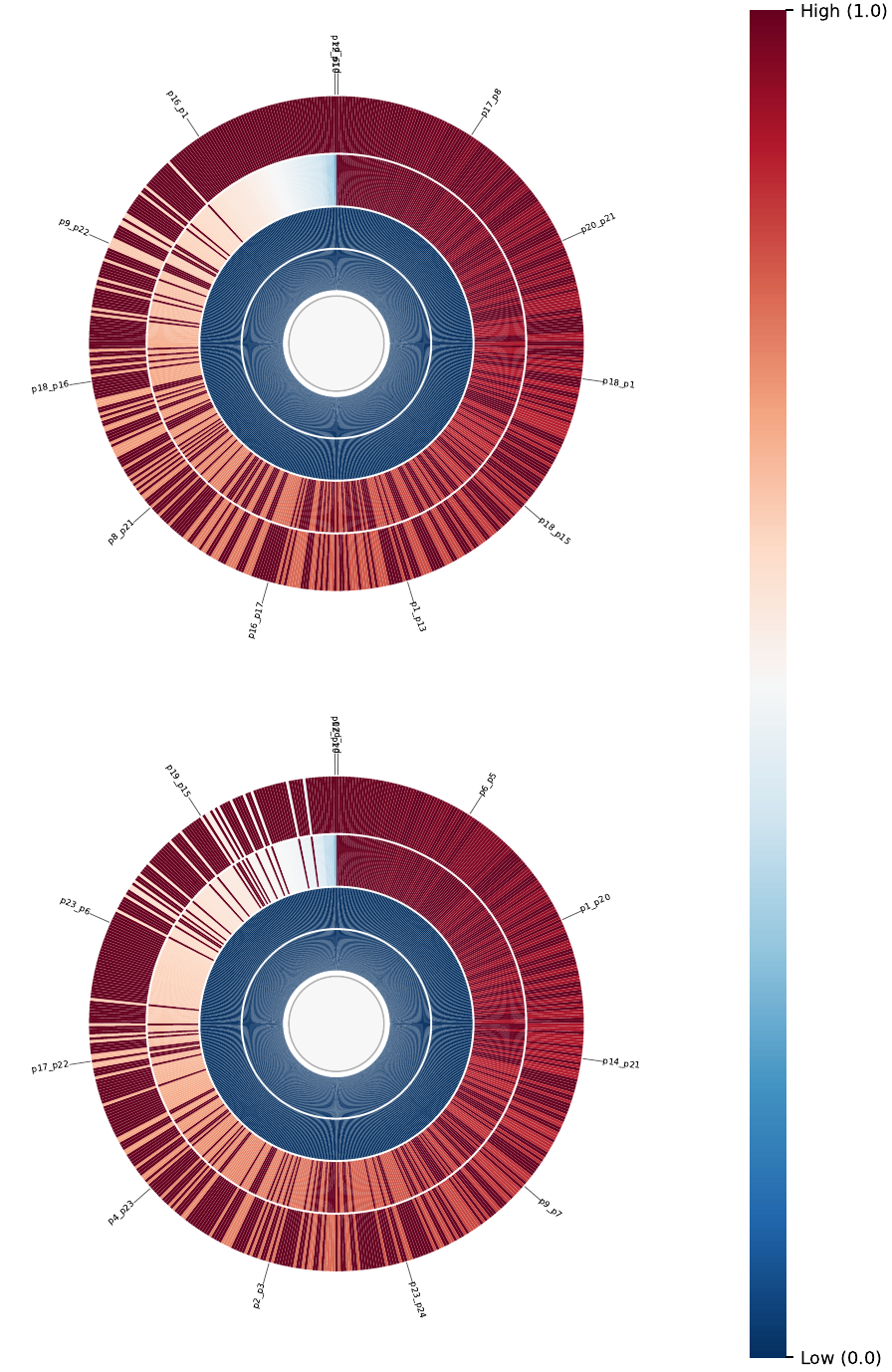}
  \caption{%
Problem-pair-level topographic maps for DE (top) and PSO (bottom), using the same representative median-seed MKKM runs as Figure~\ref{fig:sh_cluster_heatmaps}.
From inner to outer ring, the four views are DeepELA, DoE2Vec, ELA, and TransOptAS.
This figure redistributes the cluster-level view-weight information in Figure~\ref{fig:sh_cluster_heatmaps} over problem-pair space and is shown for completeness.
The visible contrast is mainly driven by the ELA--TransOptAS balance of the cluster assigned to each problem pair.%
}
  \label{fig:sh_topography_main}
\end{figure}

The relative ELA--TransOptAS contribution is cluster-dependent.
On DE, the ELA contribution ranges from 0.22 to 0.54 across the eight clusters; on PSO, it ranges from 0.19 to 0.58.
Equivalently, TransOptAS ranges from 0.46 to 0.78 on DE and from 0.42 to 0.81 on PSO.
This variation suggests that the learned solution is sparse but not collapsed to a single active representation.
The topographic map later in Figure~\ref{fig:sh_topography_main} redistributes the cluster-level view-weight information in Figure~\ref{fig:sh_cluster_heatmaps} over problem-pair space and is shown for completeness rather than as an additional quantitative result.
\label{sec:routing}

\begin{figure}[t]
\centering
\includegraphics[width=0.48\textwidth]{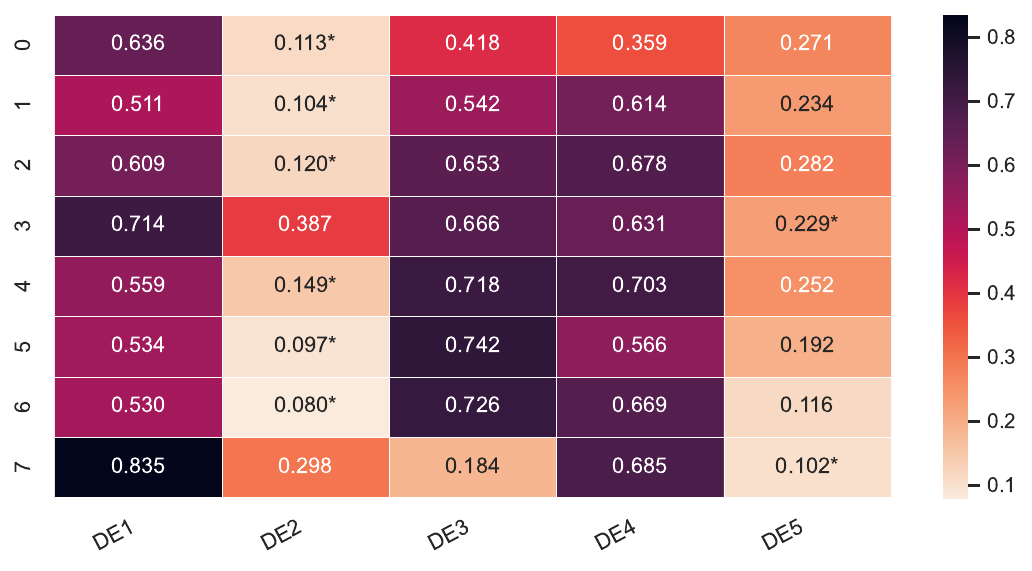}\hfill
\includegraphics[width=0.48\textwidth]{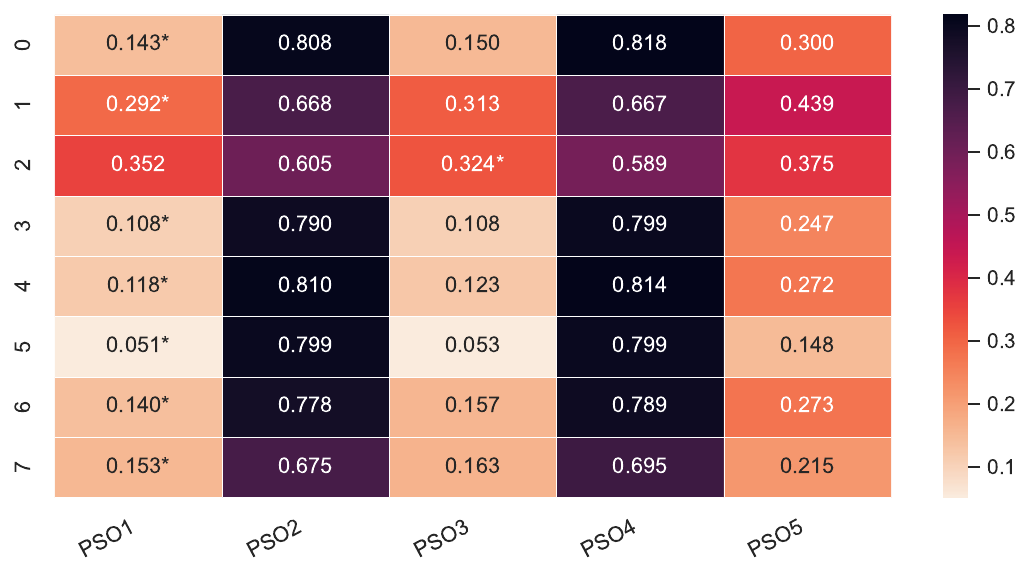}
\caption{%
Cluster-to-solver routing for DE (left) and PSO (right) in the same representative median-seed runs as Figure~\ref{fig:sh_cluster_heatmaps}.
Lower cost is better; row-wise best solvers are marked~($*$).
Patterns are empirical regularities, not intrinsic solver classes.%
}
\label{fig:sh_solver_maps}
\end{figure}

Figure~\ref{fig:sh_solver_maps} provides a post-hoc interpretability check by examining how the unsupervised clusters are routed to solver recommendations.
On DE, the routing remains clearly non-degenerate at $k=8$: clusters~0, 1, 2, 4, 5, and~6 route to DE2, whereas clusters~3 and~7 route to DE5.
The best cluster-level mean costs range from 0.080 to 0.229, and several clusters show clear separation between the best and second-best solver.
For example, cluster~2 routes to DE2 with mean cost 0.120, while the next solver among the top three has mean cost 0.282.

On PSO, the routing is more compressed but still structured.
Clusters~0, 1, 3, 4, 5, 6, and~7 route to PSO1, whereas cluster~2 routes to PSO3.
The best-versus-second-best margins are much smaller than on DE; in some clusters, such as clusters~3 and~5, PSO1 and PSO3 are almost tied.
This compressed routing is consistent with the small selector gains reported in Table~\ref{tab:main}.

A quantitative contrast reinforces this interpretation.
From the row-wise costs in Figure~\ref{fig:sh_solver_maps}, the average gap between the best and second-best solver among the top three solvers is about 0.116 on DE, but only about 0.011 on PSO.
We therefore interpret the larger DE gains and the marginal PSO gains as consequences of different solver-response separability under the present benchmark, portfolio, and EC cost criterion.

\clearpage
\clearpage

\subsection{Cluster Granularity Sensitivity}
\label{sec:k_sweep}

\begin{figure}[htbp]
  \centering
  \includegraphics[width=1\textwidth]{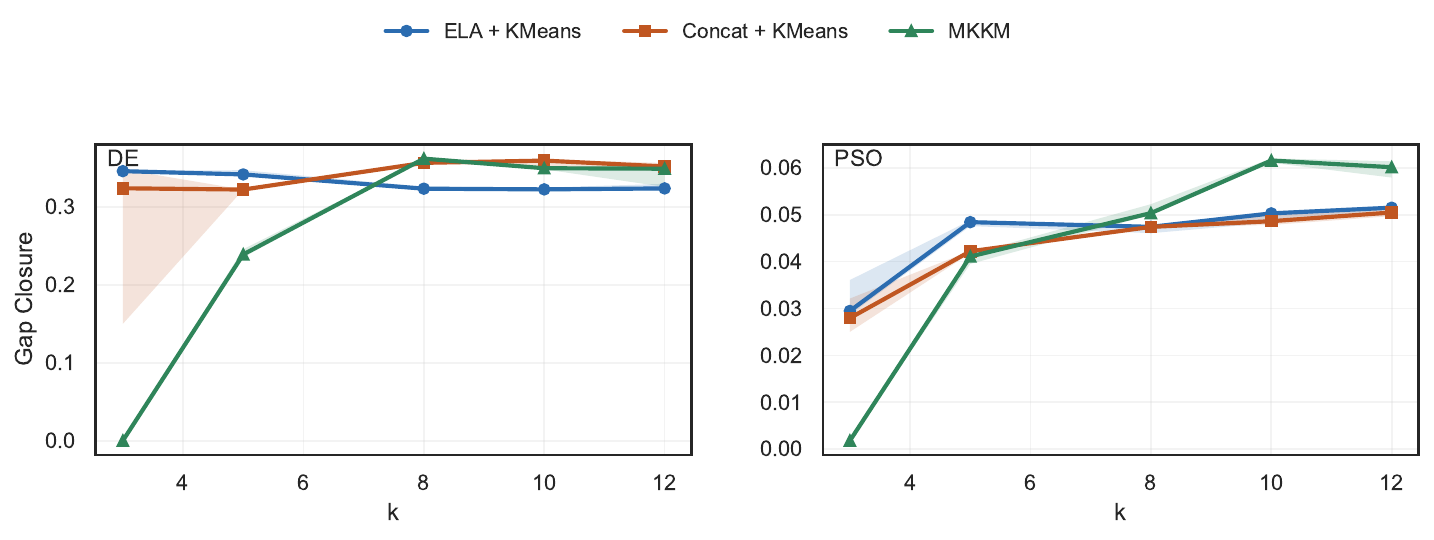}
  \caption{%
Cluster-granularity diagnostic over $k\in\{3,5,8,10,12\}$ for DE and PSO, evaluated by Gap Closure.
Curves summarize 50-seed stochastic profiles for the compared configurations.
Table~\ref{tab:main} reports the fixed $k=8$ comparison; the other $k$ values are included only as a granularity-sensitivity diagnostic.%
}
  \label{fig:k_sweep}
\end{figure}

Figure~\ref{fig:k_sweep} summarizes a diagnostic sweep over $k\in\{3,5,8,10,12\}$ for MKKM and two simple diagnostic baselines, Concat+$k$-Means and ELA+$k$-Means.
This sweep assesses the sensitivity of selector utility to the clustering granularity. Table~\ref{tab:main} reports the fixed $k=8$ comparison, where $k=8$ is used as a moderate over-partition of the five-solver portfolios.
For $k<|\mathcal{A}|=5$, the selector can use at most $k$ distinct solvers, so those points are structurally constrained.
For $k>|\mathcal{A}|$, multiple clusters may route to the same solver and the extra clusters act as a finer partition of the problem space.

On DE, $k=8$ lies on a competitive plateau for MKKM rather than being an isolated optimum.
The sweep also shows that Concat+$k$-Means remains strong for nearby values of $k$, supporting the cautious interpretation that MKKM improves over a strong concatenation baseline rather than a weak one.
On PSO, the selector landscape is compressed at $k=8$, and larger granularities may further improve MKKM.
We therefore treat the larger-$k$ PSO results as evidence that the role of $k$ as a resolution parameter is family-specific, and as a direction for future tuning rather than as the central claim of this study.

\label{sec:sensitivity}

\begin{figure}[htbp]
  \centering
  \includegraphics[width=0.5\textwidth]{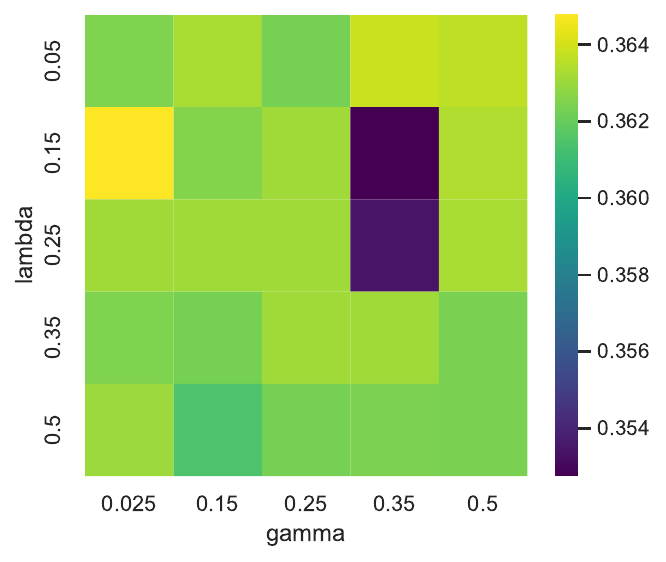}\hfill
  \includegraphics[width=0.5\textwidth]{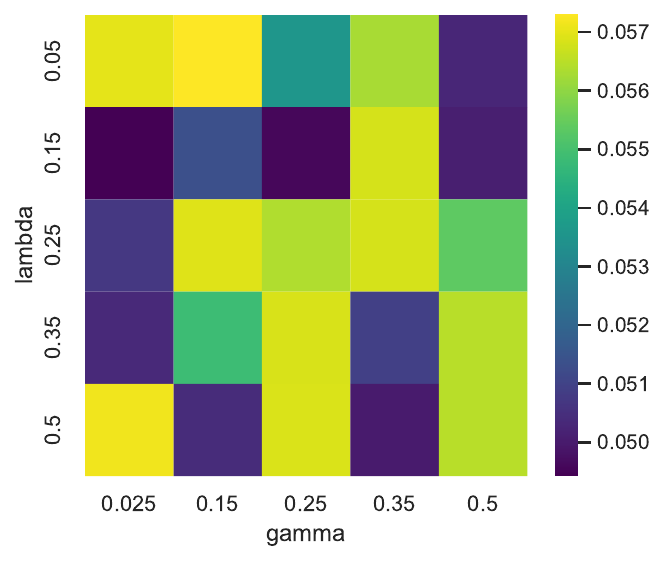}
  \caption{%
Sensitivity of multi-kernel clustering to $(\lambda,\gamma)$ on DE (left) and PSO (right) at $k=8$, evaluated by Gap Closure.
The sweep shows broad competitive regions rather than a single isolated optimum.%
}
  \label{fig:sensitivity}
\end{figure}

Figure~\ref{fig:sensitivity} reports a compact sensitivity sweep around the sparse, non-collapsed regularization region used in the main experiments at $k=8$.
The heatmaps summarize median Gap Closure over the stochastic seeds for each $(\lambda,\gamma)$ cell.
On DE, the median GC values in the explored grid are concentrated around the 0.36 level, with the best cells near 0.365.
On PSO, the absolute range is narrower in practical terms, with the leading cells around 0.057 and many cells remaining close to the 0.05 level.

The near-flat EC metric surfaces do not imply that the clustering objective itself is insensitive to $(\lambda,\gamma)$.
Rather, within the explored range, different unsupervised solutions can induce similar post-hoc routing decisions and therefore similar selector-level metrics.
The practical implication is that the method does not require a highly precise regularization setting to achieve competitive selector utility in the present benchmark.
At the same time, the PSO surface and the $k$-sweep both caution against overclaiming: small absolute GC differences in this family should be interpreted as weak evidence unless confirmed across broader benchmark and portfolio settings.

\section{Conclusion and Discussion}
\label{sec:conclusion}

We introduced unsupervised multi-kernel learning as a principled framework for automated algorithm selection from heterogeneous landscape representations.
By clustering problem instances directly from landscape characteristics and treating each representation as a separate kernel source, the approach preserves view-specific geometry and avoids forcing heterogeneous views into a single concatenated feature space.
The experiments report mean$\pm$standard-deviation selector profiles over 50 independent random seeds at the fixed evaluation budget and $k=8$.

On the affine BBOB-derived selector tasks studied here, MKKM obtains the strongest mean profile among the compared $k=8$ configurations on DE, with GC=$0.3605\pm0.0042$, MR=$0.0524\pm0.0003$, and SC=$0.1188\pm0.0003$.
This advantage is clear relative to the strongest single-view mean baseline and modest relative to the strongest concatenation mean baseline.
On PSO, MKKM is also nominally ahead by mean GC, MR, and SC, but the margin over the strongest concatenation baseline is very small relative to the stochastic standard deviation and should not be interpreted as strong dominance.
We therefore interpret the PSO result as competitive and mildly favorable rather than as strong dominance.

The learned kernel weights in representative median-seed runs retain ELA and TransOptAS while assigning zero weight to DeepELA and DoE2Vec.
This should be interpreted conditionally on the present benchmark, fixed budget, portfolio, and regularization regime, not as a universal statement about the value of these representations.
The routing analysis further indicates markedly stronger solver differentiation on DE than on PSO, with an average best-versus-second-best cluster-level gap of approximately 0.116 on DE versus 0.011 on PSO.
Together, these results suggest that multi-kernel clustering is most useful when the learned clusters separate solver-response regimes with sufficiently different cost profiles.

Several directions follow naturally from this work.
It will be important to test the robustness of these findings across other benchmarks, problem dimensions, budgets, and portfolio compositions, and to examine richer kernel families that may capture nonlinear structure beyond the present setting.
At the decision level, softer or uncertainty-aware cluster-to-solver mappings may reduce the gap to the Virtual Best Solver, while evaluations on real-world optimization problems and mixed search spaces will be necessary to assess practical generalization.
A particularly promising direction is to combine the present unsupervised grouping stage with lightweight supervised refinement using only a small amount of labeled data, thereby exploring a middle ground between fully unsupervised selection and label-intensive supervised AAS.
Overall, these results indicate that multi-kernel learning provides a competitive, interpretable, and label-free foundation for unsupervised algorithm selection, while also motivating further work on evaluation principles that align unsupervised structure discovery with downstream selector utility.

\subsubsection{\ackname}
Y.~Lu acknowledges the support from Marie Skłodowska-Curie Grant No.~101081674 (SOUND.AI) of the European Union’s Horizon Europe research and innovation programme. T.~Eftimov acknowledges the support of Horizon Europe ERA Chair AutoLearn-SI (101187010), as well as the Slovenian Research Agency through programme grant No.~P2-0098 and project grants No.~J2-70078 and No.~GC-0001. C.~Doerr acknowledges funding by the European Union (ERC, ``dynaBBO'', grant no.~101125586). Views and opinions expressed are however those of the author(s) only and do not necessarily reflect those of the European Union or the European Research Council Executive Agency. Neither the European Union nor the granting authority can be held responsible for them.
This research was also jointly funded by the French National Research Agency (ANR-23-CE23-0035) and the German Research Foundation (DFG; LI 2801/7-1), through project \textsc{Opt4DAC}.

\newpage

\end{document}